\documentclass[final, 5p, authoryear, lefttitle, sort&compress]{elsarticle}

\usepackage{soul}
\usepackage[ruled,vlined]{algorithm2e}
\usepackage{mathtools}
\usepackage{subfigure}
\usepackage{xcolor}

\usepackage{array}
\usepackage{bookmark}
\usepackage{graphicx}
\usepackage{url}   
\usepackage{amsmath}

\usepackage{lineno,hyperref}
\modulolinenumbers[5]
\usepackage{etoolbox}
\patchcmd{\pprintMaketitle}
{\ifvoid\absbox\else\unvbox\absbox\par\vskip10pt\fi}
{\ifvoid\absbox\else\clearpage\unvbox\absbox\par\vskip30pt\fi}
{}{}
\patchcmd{\pprintMaketitle}
{\hrule\vskip12pt}
{}
{}{}
\patchcmd{\pprintMaketitle}
{\hrule\vskip12pt}
{}
{}{}
\appto{\pprintMaketitle}{\clearpage}

\begin{document}

\begin{frontmatter}

\title{Refined Continuous Control of DDPG Actors via Parametrised Activation\corref{lic}}

\author[chpl]{Mohammed~Hossny\corref{corr}}
\cortext[corr]{Corresponding author}
\cortext[lic]{\textcopyright 2020. This manuscript version is made available under the CC-BY-NC-ND 4.0 license http://creativecommons.org/licenses/by-nc-nd/4.0/}

\author[deakin]{Julie~Iskander}
\author[alex]{Mohammed Attia}
\author[uts]{Khaled Saleh}
\address[deakin]{Deakin University, Australia}
\address[alex]{Medical Research Institute, Alexandria University, Egypt}

\address[uts]{University of Technology Sydney, Australia}

\begin{abstract}
\textbf{Introduction:}
Continuous action spaces impose a serious challenge for reinforcement learning agents. While several off-policy reinforcement learning algorithms provide a universal solution to continuous control problems, the real challenge lies in the fact that different actuators feature different response functions due to wear and tear (in mechanical systems) and fatigue (in biomechanical systems). 

\smallbreak
\noindent\textbf{Contribution:}
In this paper, we propose enhancing actor-critic reinforcement learning agents by parameterising the final actor layer which produces the actions in order to accommodate the behaviour discrepancy of different actuators, under different load conditions during interaction with the environment. 

\smallbreak
\noindent\textbf{Methods:}
We propose branching the action producing layer in the actor to learn the tuning parameter controlling the activation layer (e.g. Tanh and Sigmoid). The learned parameters are then used to create tailored activation functions for each actuator. 

\smallbreak
\noindent\textbf{Results:}
We ran experiments on three OpenAI Gym environments, i.e. \texttt{Pendulum-v0}, \texttt{LunarLanderContinuous-v2} and \texttt{BipedalWalker-v2}. Results have shown an average of 23.15\% and 33.80\% increase in total episode reward of the \texttt{LunarLanderContinuous-v2} and \texttt{BipedalWalker-v2} environments, respectively. There was no significant improvement in \texttt{Pendulum-v0} environment but the proposed method produces a more stable actuation signal compared to the state-of-the-art method. 


\smallbreak
\noindent\textbf{Significance:}
The proposed method allows the reinforcement learning actor to produce more robust actions that accommodate the discrepancy in the actuators' response functions. This is particularly useful for real life scenarios where actuators exhibit different response functions depending on the load and the interaction with the environment. This also simplifies the transfer learning problem by fine tuning the parameterised activation layers instead of retraining the entire policy every time an actuator is replaced. Finally, the proposed method would allow better accommodation to biological actuators (e.g. muscles) in biomechanical systems.


\end{abstract}

\begin{keyword}
\texttt{Continuous Control}\sep \texttt{Deep Reinforcement Learning}\sep \texttt{Actor-Critic}\sep \texttt{DDPG}
\end{keyword}

\end{frontmatter}

\section{Introduction}
\label{sc:intro}

Deep reinforcement learning (DRL) has been used in different domains; and has achieved good results on different tasks such as robotic control, natural language processing and biomechanical control of human models~\citep{kidzinski2020artificial,kidzinski2018learning, mnih2015human,kober2013reinforcement}. 

\smallbreak
While DRL is proven to handle discrete problems effectively and efficiently, continuous control remains a challenging task to accomplish. This is because it relies on physical systems which are prone to noise due to wear and tear, overheating, and altered actuator response function depending on the load each actuators bears; this is more apparent in robotic and biomechanical control problems. In the robotic control domain, for instance, Bi-pedal robots robustly performs articulated motor movements under complex environments and limited resources. These robust movements are achieved using highly sophisticated model-based controllers. However, the motor characteristics are highly dependent on the load and the interaction with the environment. Thus, adaptive continuous control is required to adapt to new situations. 

\smallbreak
Biomechanical modelling and simulation present a clearer example. In a biomechanical system, the human movement is performed using muscle models~\citep{thelen2003adjustment,millard2013flexing}. These models simulate muscle functions, which are complex and dependent on multiple parameters, like muscle maximum velocity, muscle optimal length, muscle maximum isometric force to name a few~\citep{zajac1989muscle}.

\smallbreak
The common challenge facing training DRL agents on continuous action spaces, is the flow of the gradient update throughout the network. The current state of the art is relying on a single configuration of the activation function producing the actuation signals. However, different actuators exhibit different transfer functions; and also noisy feedback from the environment propagates through the entire actor neural network and thus, a drastic change is imposed on the learned policy. The solution we are proposing in this work is to use multiple actuation transfer functions that allow the actor neural network to adaptively modify the actuation response functions to the needs of each actuator.



\smallbreak
In this paper, we present a modular perspective of the actor in actor-critic DRL agents and propose modifying the actuation layer to learn the parameters defining the actuation-producing activation functions (e.g. Tanh and Sigmoid). It is important to emphasise the difference between parameterised action spaces and parameterised activation functions. In reinforcement learning, a parametrised action space is commonly referred to as a discrete action space that has an accompanying one or more continuous parameters~\citep{masson2016reinforcement}. It has been used to solve problems such as the RoboCup~\citep{kitano1997robocup}, which is a robots world-cup soccer game~\citep{hausknecht2015deep}. On the other hand, parameterised activation functions, such as PReLU~\citep{He20151026} and SeLU~\citep{klambauer2017selfnormalizing}, were introduced to combat overfitting and saturation problems. In this paper, we adopt parameterised activation functions to improve performance of the deep deterministic policy gradient (DDPG) to accommodate the complex nature of real-life scenarios. The rest of this paper is organised as follows. Related work is discussed in Section~\ref{sc:bkg}. Proposed method is presented in Section~\ref{sc:proposed}. Experiments and results are presented in Section~\ref{sc:exp_res}. Finally, Section~\ref{sc:conc} concludes and introduces future advancements.

\section{Background}
\label{sc:bkg}
Deep deterministic policy gradient (DDPG) is a widely adopted deep reinforcement learning method for continuous control problems~\citep{lillicrap2015continuous}. 
A DDPG agent relies on three main components; the actor, the critic and the experience replay buffer~\citep{lillicrap2015continuous}. 

\smallbreak
In the actor-critic approach~\citep{sutton1998introduction}, the actor neural network reads observations from the environment and produces actuation signals. After training, the actor neural network serves as the controller which allows the agent to navigate the environment safely and to perform the desired tasks. 
The critic network assesses the anticipated reward based on the current observation and the actor's action. In control terms, the critic network serves as a black-box system identification module which provides guidance for tuning the parameters of a PID controller. The observations, actions, estimated reward and next state observation are stored as an experience in a circular buffer. This buffer serves as a pool of experiences, from where samples are drawn to train the actor and the critic neural networks to produce the correct action and estimate the correct reward, respectively. 

\smallbreak
There are different DDPG variations in the literature. In~\citep{fujimoto2018addressing}, a twin delay DDPG (TD3) agent was proposed to limit overestimation by using the minimum value between a pair of critics instead of one critic. In~\citep{Barth-Maron2018}, it was proposed to expand the DDPG as a distributed process to allow better accumulation of experiences in the experience replay buffer. Other off-policy deep reinforcement learning agents such as soft actor-critic (SAC) are inspired by DDPG although they rely on stochastic parameterisation~\citep{haarnoja2018soft,haarnoja2018softarx}. In brief, SAC adapts the reparameterisation trick to learn a statistical distribution of actions from which samples are drawn based on the current state of the environment.

\subsection{DDPG Challenges}
Perhaps the most critical challenge of the DDPG, and off-policy agents in general, is its sample inefficiency. The main reason behind this challenge is that the actor is updated depending on the gradients calculated by the continuously training of the critic neural network. This gradient is noisy because it relies on the outcome of the simulated episodes. Therefore, the presence of outlier scenarios impact the training of the actor and thus constantly change the learned policy instead of refining it. This is the main reason off-policy DRL training algorithms require maintaining a copy of the actor and critic neural networks to avoid divergence during training. 

While radical changes in the learned policy may provide a good exploratory behaviour of the agent, it does come at the cost of requiring many more episodes to converge. Additionally, it is often recommended to have controllable exploratory noise parameters separated from the policy either by inducing actuation noise such as Ornstein–Uhlenbeck~\citep{UhlenbeckOrnstein} or maximising the entropy of the learned actuation distribution \citep{haarnoja2018soft,haarnoja2018softarx}. Practically, however, for very specific tasks, and most continuous control tasks are, faster convergence is often a critical aspect to consider. Another challenge, which stems from practical applications, is the fact that actuators are physical systems and are susceptible to having different characterised transfer functions in response to the supplied actuation signals. These characterisation discrepancies are almost present in every control system due to wear and tear, fatigue, overheating, and manufacturing factors. While minimal discrepancies are easily accommodated with a PID controller, they impose a serious problem with deep neural networks. This problem, in return, imposes a serious challenge during deployment and scaling operations. 

\section{Proposed Method}
\label{sc:proposed}
In order to address the aforementioned challenges, we propose parameterising the final activation function to include scaling and translation parameters $k, x_0$. In our case, we used $\tanh\left(kx-kx_0\right)$ instead of $\tanh(x)$ to allow the actor neural network to accommodate the discrepancies of the actuator characteristics by learning $k$ and $x_0$. The added learnable parameters empower the actor with two additional degrees of freedom. 

\begin{figure}
    \centering
    \includegraphics[width=.96\linewidth]{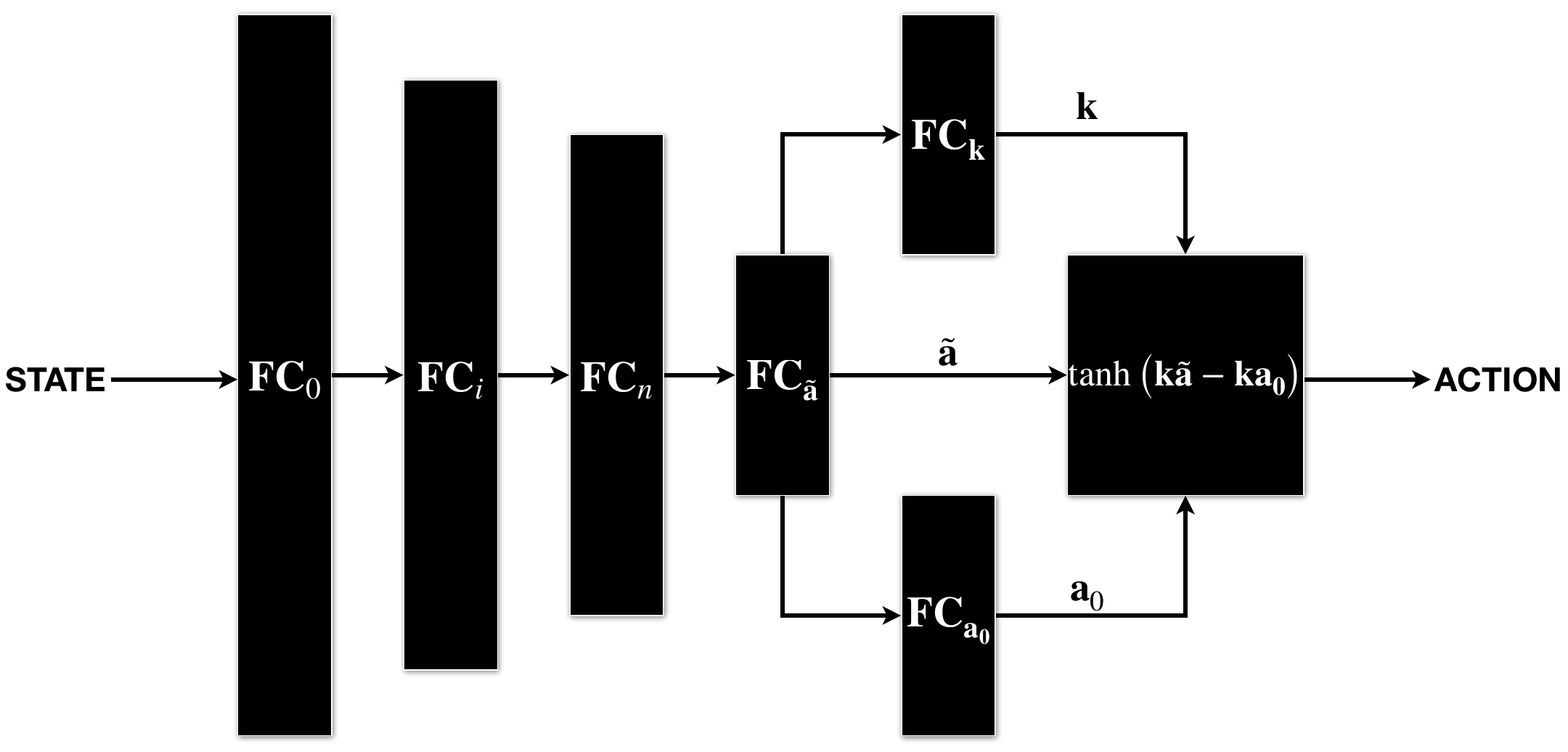}

    \caption{Proposed modification to the actor. The final fully connected layer branches into two fully connected layers to learn the $x_0$ and $k$ parameters of $\tanh\left(kx-kx_0\right)$.}
    \label{fig:prmTanhModel}
\end{figure}

\subsection{Modular Formulation}
In a typical DRL agent, an actor consists of several neural network layers. While the weights of all layers collectively serve as a policy, they do serve different purposes based on their interaction with the environment. The first layer encode observations from the environment and thus we propose to call it the observer layer. The encoded observations are then fed into several subsequent layers and thus we call them the policy layers. Finally, the output of the policy layers are usually fed to a single activation function. Throughout this paper, we will denote to the observer, policy and action parts of the policy neural network as $\pi^O$, $\pi^P$, $\pi^A$, respectively. We will also denote to the observation, the pre-mapped action space, and the final action space as $O$, $\tilde A$ and $A$, respectively. To that end, the data flow of the observation $o_t \in O$ through the policy $\pi$ to produce an action $a_t \in A$ can be summarised as;
\begin{eqnarray}
a_t &=& \pi(o_t)= \pi^A\circ\pi^P\circ\pi^O\left(o_t\right),\\
~  &=& \pi^A\left(\pi^P\left(\pi^O\left(o_t\right)\right)\right), \label{eq:a_t}
\end{eqnarray}
where $\pi^O: O \rightarrow \tilde A$, $\pi^P: \tilde O \rightarrow \tilde A$ and $\pi^A: \tilde A \rightarrow A$. 

\smallbreak
In a typical actor, there is no distinction between the observer and policy layers. Also, the actuation layer is simply regarded as the final activation function $\pi^A(x)=\tanh(x)$ and thus the actor is typically modelled as one multi layer perceptron neural network (MLP). The problem with having $\pi^A$ as $\tanh$\footnote{Sigmoid is also a popular activation function where $A=[0, 1]$.} is that it assumes that  all actuators in the system exhibit the same actuation-torque characterisation curves under all conditions.

\begin{figure}[t]
\centering
\includegraphics[trim={0mm 0 6mm 0},clip,width=1\linewidth]{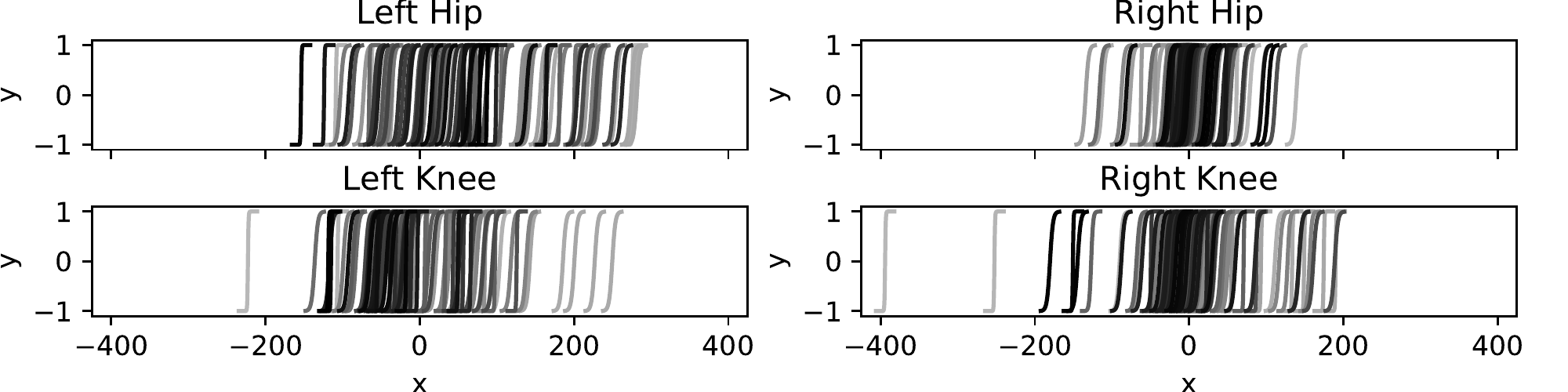}
\caption{Desired parameterisation of the $\tanh\left(kx-kx_0\right)$ activation function. Allowing extra degrees of freedom empowers the actor neural network to accommodate outlier scenarios with minimal update to the actual policy. The results here are from the proposed actor trained and tested on bipedal walker environment. Colour brightness indicate different stages throughout the episode from start (bright) to finish (dark).}
\label{fig:res_tanh}
\end{figure}

\subsection{Parameterising $\pi^A$}
Beccause actuation characterisation curves differ based on their role and interaction with the environment, using a single activation function, forces the feedback provided by the environment to propagate throughout the gradients of the entire policy. Therefore, we chose to use a parameterised $\pi^A\left(kx-kx_0\right)$ to model the scaling and the translation of the activation function, and thus the data flow in Eq.~\ref{eq:a_t} can be expanded as;
\begin{eqnarray}
k_t  &=& \pi^K\left(\pi^P\left(\pi^O\left(o_t\right)\right)\right),\\
a^0_t  &=& \pi^{a_0}\left(\pi^P\left(\pi^O\left(o_t\right)\right)\right),\\
a_t  &=& \pi^A\left(k_t\pi^P\left(\pi^O\left(o_t\right)\right)-k_t a^0_t\right), 
\end{eqnarray}
where $\pi^{a0}$, $\pi^k$ are simple fully connected layers and $\pi^A$ remains an activation function (i.e. $\tanh$) as shown in Fig.~\ref{fig:prmTanhModel}. Adjusting the activation curves based on the interaction with the environment allows the policy to remain intact and thus leads to a more stable training as discussed in the following section. Figure~\ref{fig:res_tanh} shows the learned parameterised $\tanh$ activation functions of the bipedal walker problem.

\smallbreak
While the automatic differentiation engines are capable of adjusting the flow of gradient updates, there are two implementation considerations to factor in the design of the actor. 
First, the scale degree of freedom parameterised by $k$, in the case of $\tanh$ and sigmoid does affect the slope of the activation function. A very small $k<0.1$ will render $\pi^A$ to be almost constant while a very high $k>25$ produces a square signal. Both extreme cases impose problems to the gradient calculations. On one hand, a constant signal produces zero gradients and prevents the policy neural network from learning. On the other hand, a square signal produces unstable exploding gradients. Another problem also occurs when $k<0$, which usually changes the behaviour of the produced signals. Therefore, we recommend using a bounded activation function after $\pi^k$ when estimating $k_t$. 

\smallbreak
Second, the translation degree of freedom parameterised by $a^0$, allows translating the activation function to an acceptable range which prevents saturation. However, this may, at least theoretically, allow the gradients of the policy $\pi^P$ and observer $\pi^O$ layers to have monotonically increasing gradients as long as the $a^0_t$ can accommodate. This in return may cause an exploding gradient problem. In order to prevent this problem we recommend employing weight normalisation  after calculating the gradients~\citep{Salimans2016901}.










\begin{figure}
    \centering
    \subfigure[Inv. Pendulum]{
    \setlength{\fboxsep}{0pt}%
    \setlength{\fboxrule}{.25pt}%
        \fbox{\includegraphics[width=.31\linewidth]{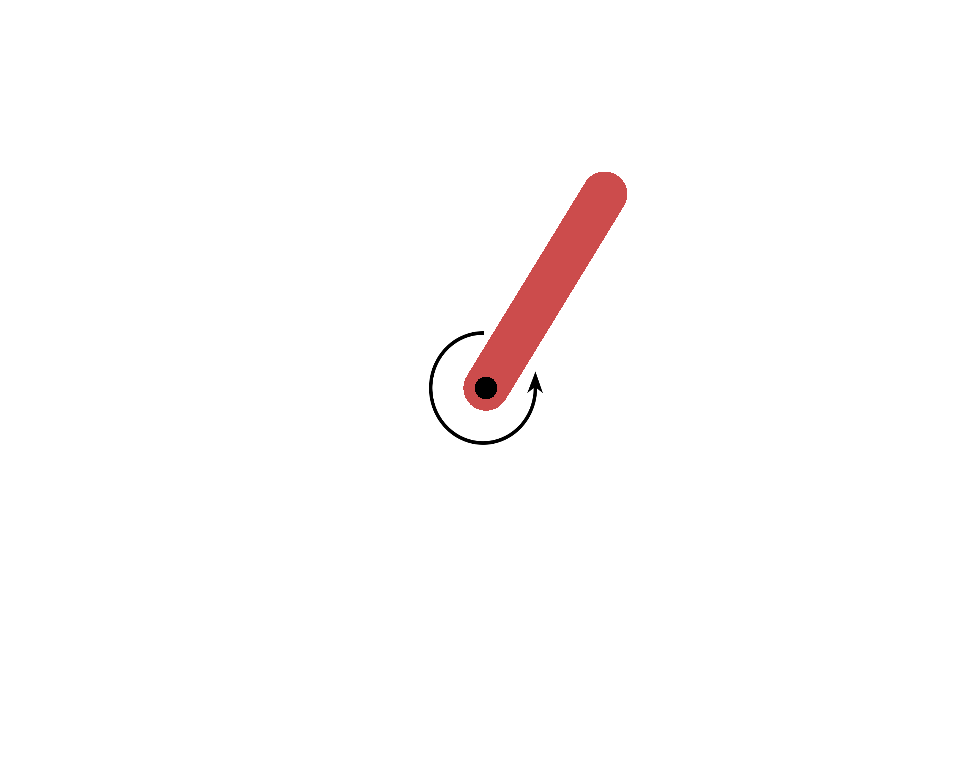}}
    }%
    \subfigure[Lunar Lander]{
    \setlength{\fboxsep}{0pt}%
    \setlength{\fboxrule}{.25pt}%
        \fbox{\includegraphics[width=.31\linewidth]{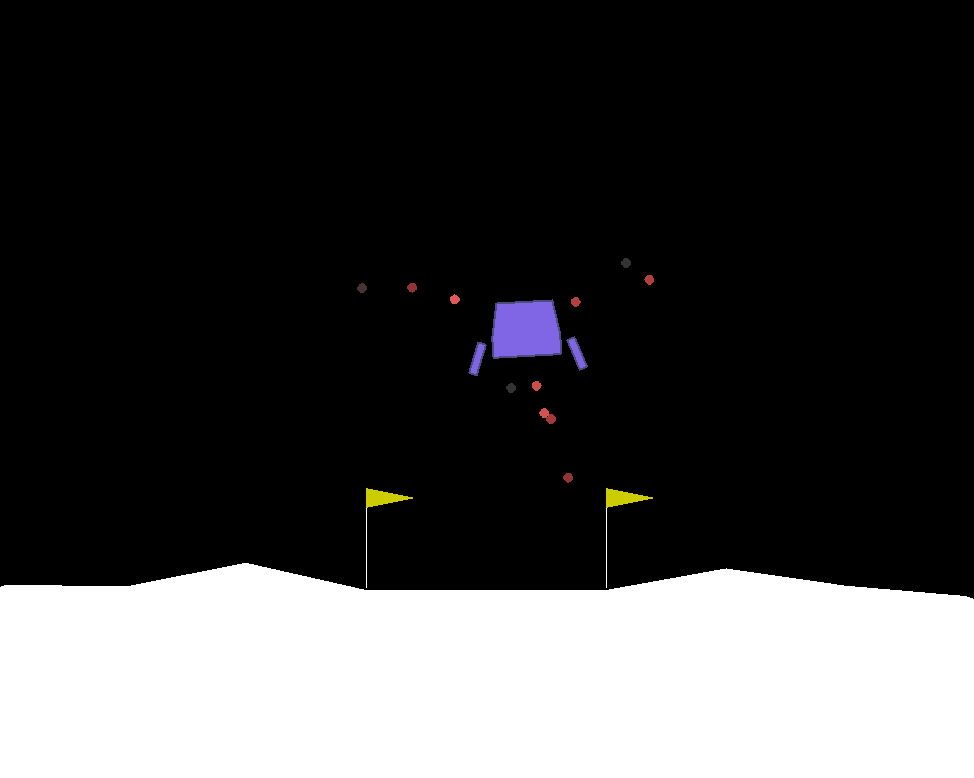}}
    }%
    \subfigure[Bipedal Walk]{
    \setlength{\fboxsep}{0pt}%
    \setlength{\fboxrule}{.25pt}%
        \fbox{\includegraphics[width=.31\linewidth]{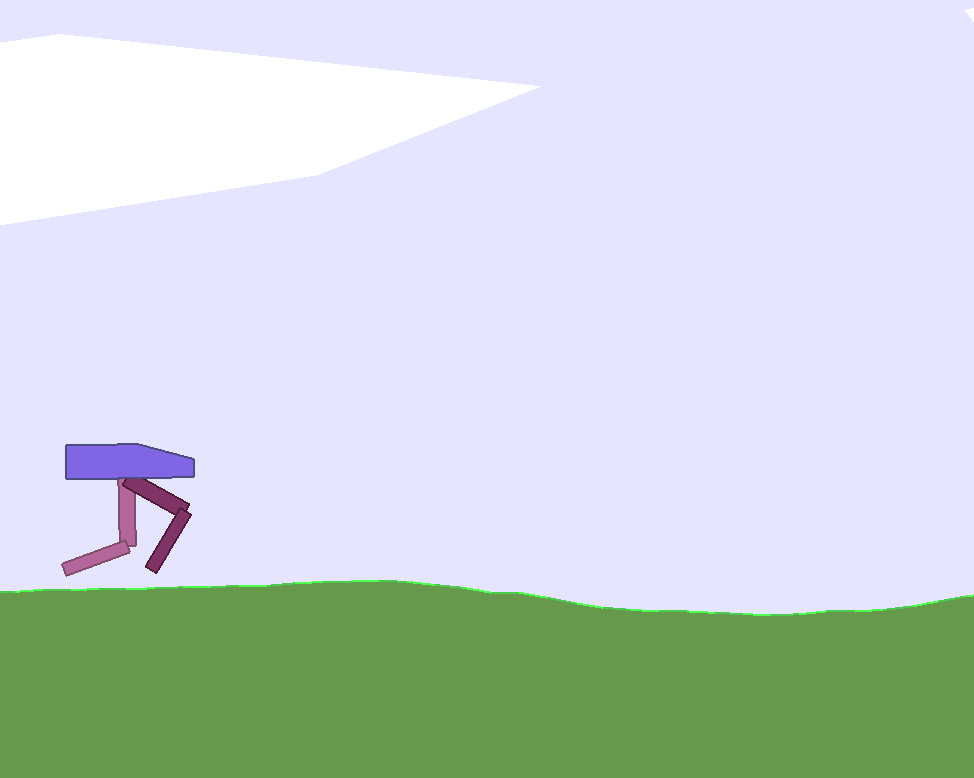}}
    }
    \caption{OpenAI gym environments used for testing.}
    \label{fig:envs}
\end{figure}

\section{Experiments and Results}
\label{sc:exp_res}
In order to test the efficacy and stability of the proposed method we trained a DDPG agent with and without the proposed learnable activation parameterisation. Both models were trained and tested on three OpenAI gym environments, shown in Fig.~\ref{fig:envs}, that are \texttt{Pendulum-v0}, \texttt{LunarLanderContinous-v2} and \texttt{BipedalWalker-v2}. For each environment six checkpoint models were saved (best model for each seed). The saved models were then tested for 20 trials with new random seeds (10 episodes with 500 steps each). The total number of test episodes is 1200 for each environment. The results of the three environments are tabulated in Tab.~\ref{tab:epsdrwrd} and Tab.~\ref{tab:steprwrd}.

\subsection{Models and Hyperparameters}
The action mapping network is where the proposed and classical models differ. The proposed model branches the final layer of into two parallel fully connected layers to infer the parameters of $k, x_0$ in $\tanh(kx-kx_0)$ activation function. The classical model adds two more fully connected layers separated by $\tanh$ activation function. The added layers ensures that the number of learnable parameters is the same in both models to guarantee a fair comparison. 

\smallbreak
Both models were trained on the three environments for the same number of episodes (200 steps each). However, number of steps may vary depending on early termination cases. The models were trained with $5$ different pseudo-random number generator (PRNG) seeds. We set the experience replay buffer to $10^6$ samples. We chose ADAM optimiser for the back-propagation optimisation and set the learning rate of both the actor and the critic to 1E-3 with first and second moments set to $0.9$, $0.999$, respectively. We set the reward discount $\gamma=0.99$ and the soft update of the target neural networks $\tau=0.005$. We also added a simple Gaussian noise with $\sigma=0.25$ to allow exploration. During the training we saved the best model (i.e. checkpoint). DDPG hyper-parameters tuning is thoroughly articulated in~\citep{lillicrap2015continuous}.







\begin{table*}[]
    \centering
    \caption{Episode Reward (mean$\pm$std). Higher mean is better.}
    \begin{tabular}{lrrr}
    \hline
    {}                   &            \texttt{Pendulum-v0} &           \texttt{LunarLanderContinuous-v2} &           \texttt{BipedalWalker-v2} \\
    \hline
                    Classic Tanh &  -269.20 $\pm$ 167.43 &  114.25 $\pm$ 41.20 &  125.27 $\pm$ 15.78 \\
       Learnable Tanh (proposed) &  -268.39 $\pm$ 166.32 &  \textbf{140.69} $\pm$ 45.21 &  \textbf{167.60} $\pm$ ${~}$8.45 \\
                          Improvement &    0.30\% &   23.15\% &   33.80\% \\
    \hline
    \end{tabular}
    \label{tab:epsdrwrd}
\end{table*}


\begin{table*}[]
    \centering
    \caption{Step Reward (mean$\pm$std). Higher mean is better.}
    \begin{tabular}{lrrr}
    \hline
    {}                   &            \texttt{Pendulum-v0} &           \texttt{LunarLanderContinuous-v2} &           \texttt{BipedalWalker-v2} \\
    \hline
                   Classic Tanh &   -0.54 $\pm$ 0.34 &  0.40 $\pm$  0.28 &    0.23 $\pm$ 0.09  \\
      Learnable Tanh (proposed) &   -0.54 $\pm$ 0.34 &  \textbf{0.67} $\pm$  0.35 &    \textbf{0.29} $\pm$ 0.16  \\
                         Improvement &   0.26\% & 65.59\% &  26.76\% \\
    \hline
    \end{tabular}
    \label{tab:steprwrd}
\end{table*}

\subsection{Inverted Pendulum Results}
\begin{figure}
    \centering
    \subfigure[Training Performance]{
    \centering
        \includegraphics[width=.96\linewidth]{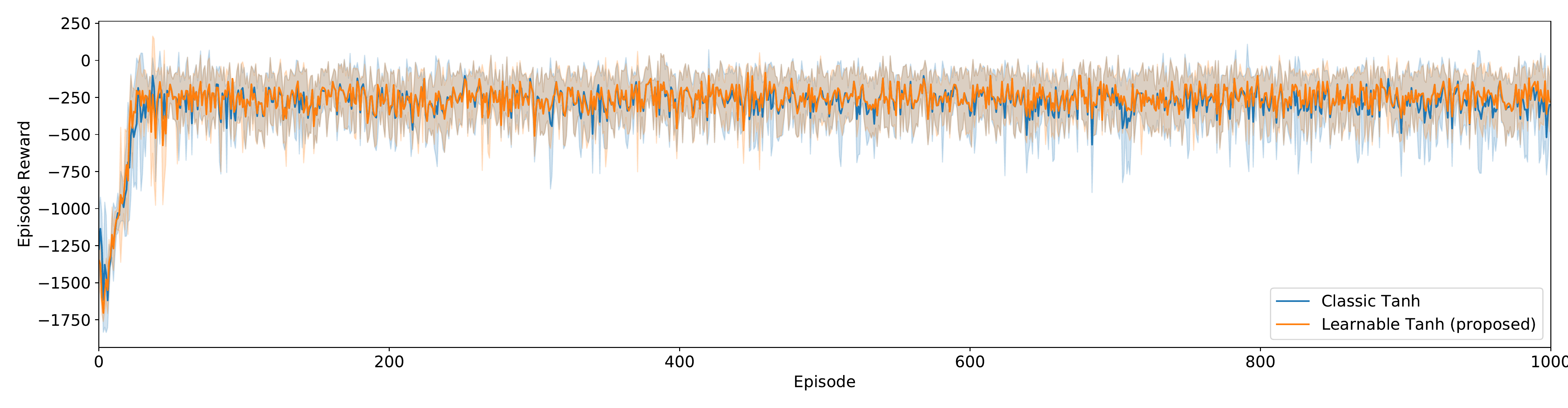}
    }\\
        \subfigure[Step Reward]{
        \label{fig:inv_stpR}
        \centering
            \includegraphics[width=.48\linewidth]{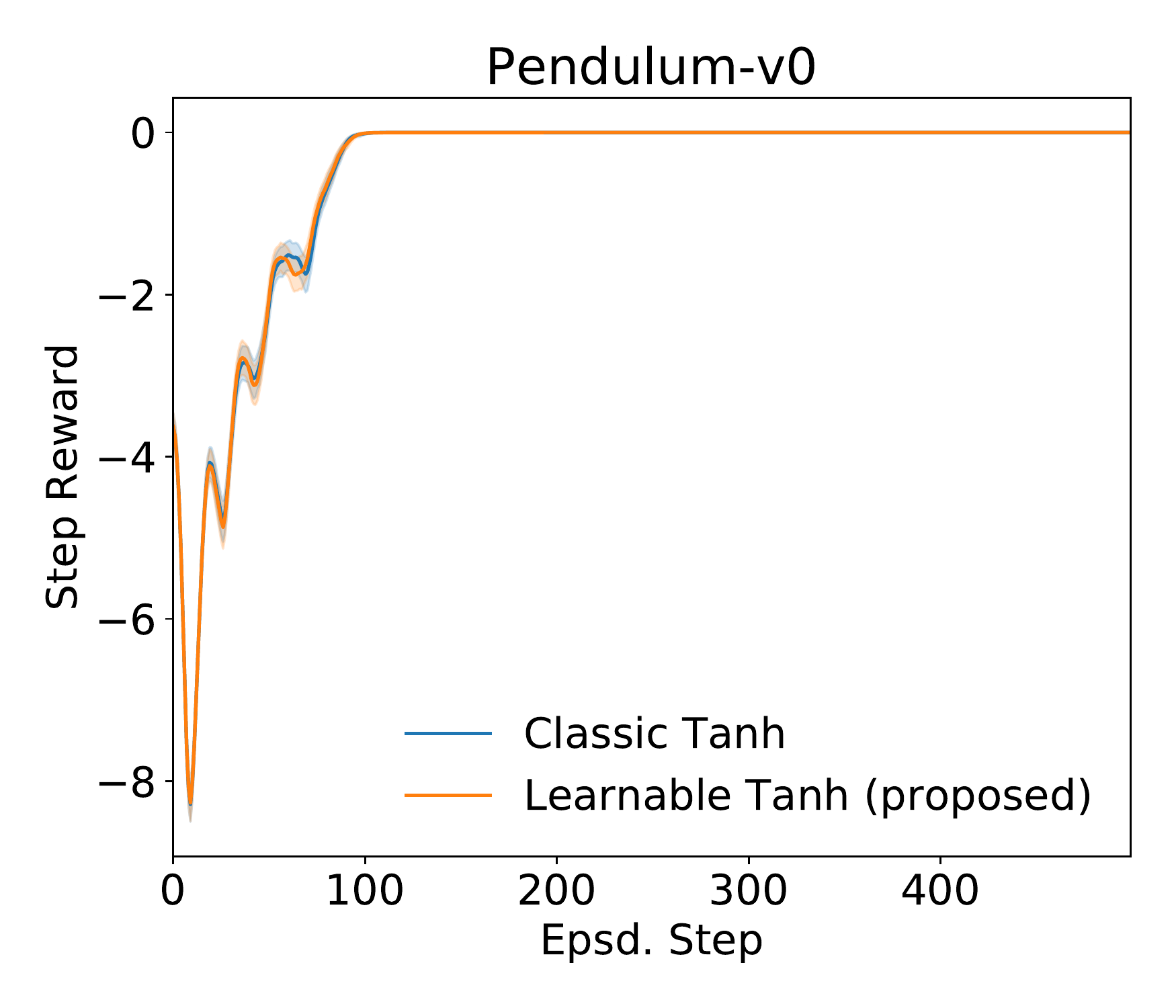}}
        \subfigure[Epsd. Reward]{
        \label{fig:inv_epsR}
        \centering
            \includegraphics[width=.48\linewidth]{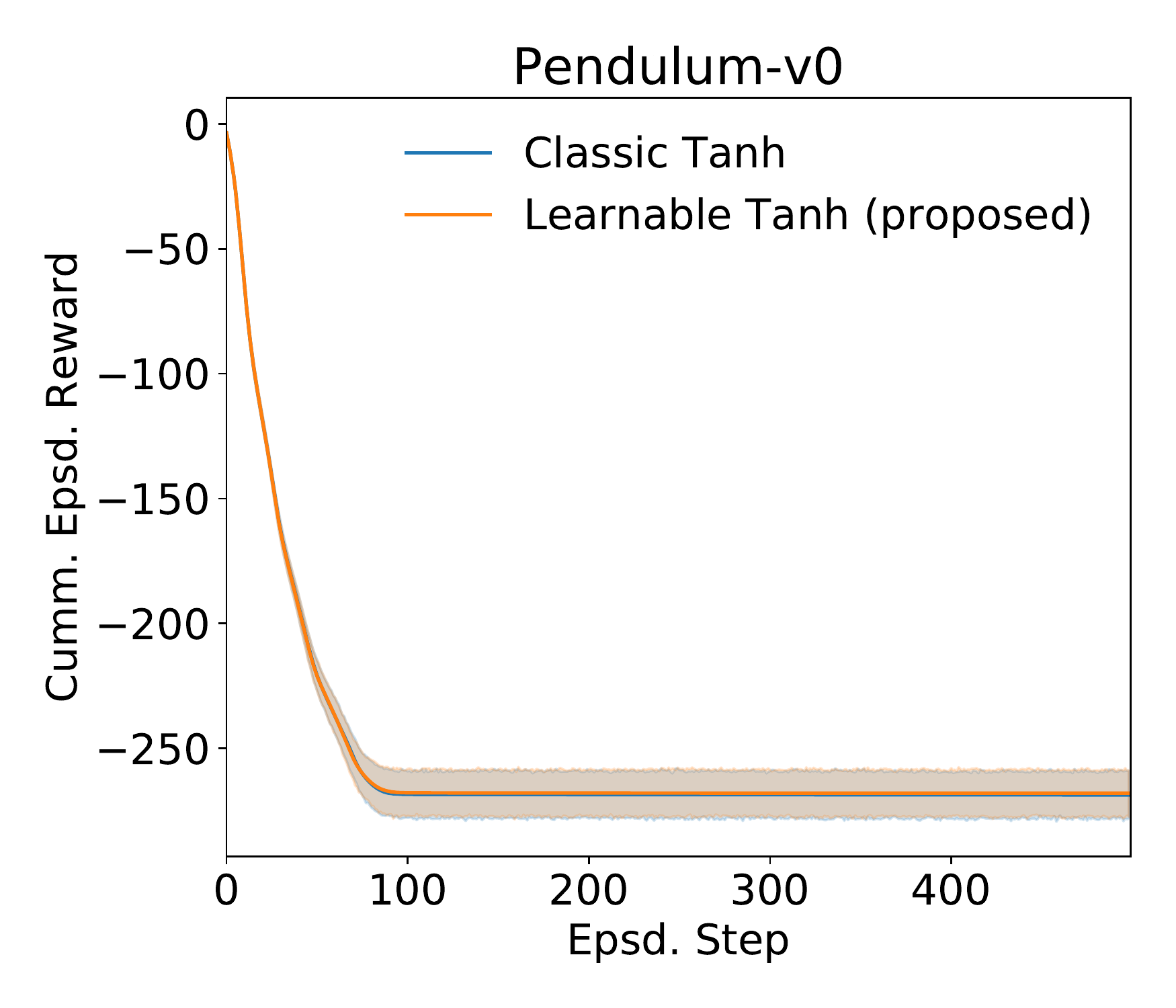}}
        \subfigure[Observation]{
        \label{fig:inv_obs}
        \centering
            \includegraphics[width=.48\linewidth]{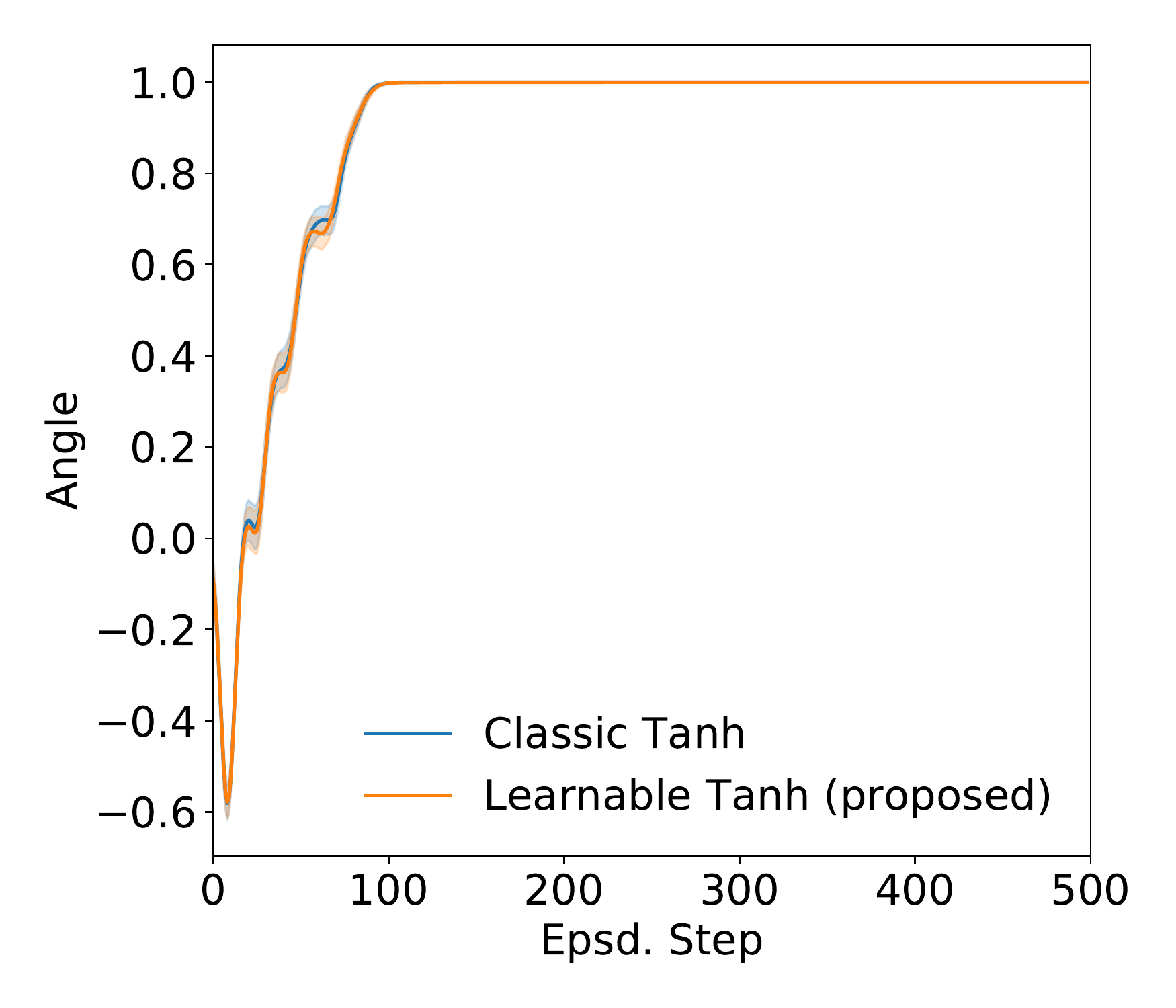}}
        \subfigure[Action]{
        \label{fig:inv_act}
        \centering
            \includegraphics[width=.48\linewidth]{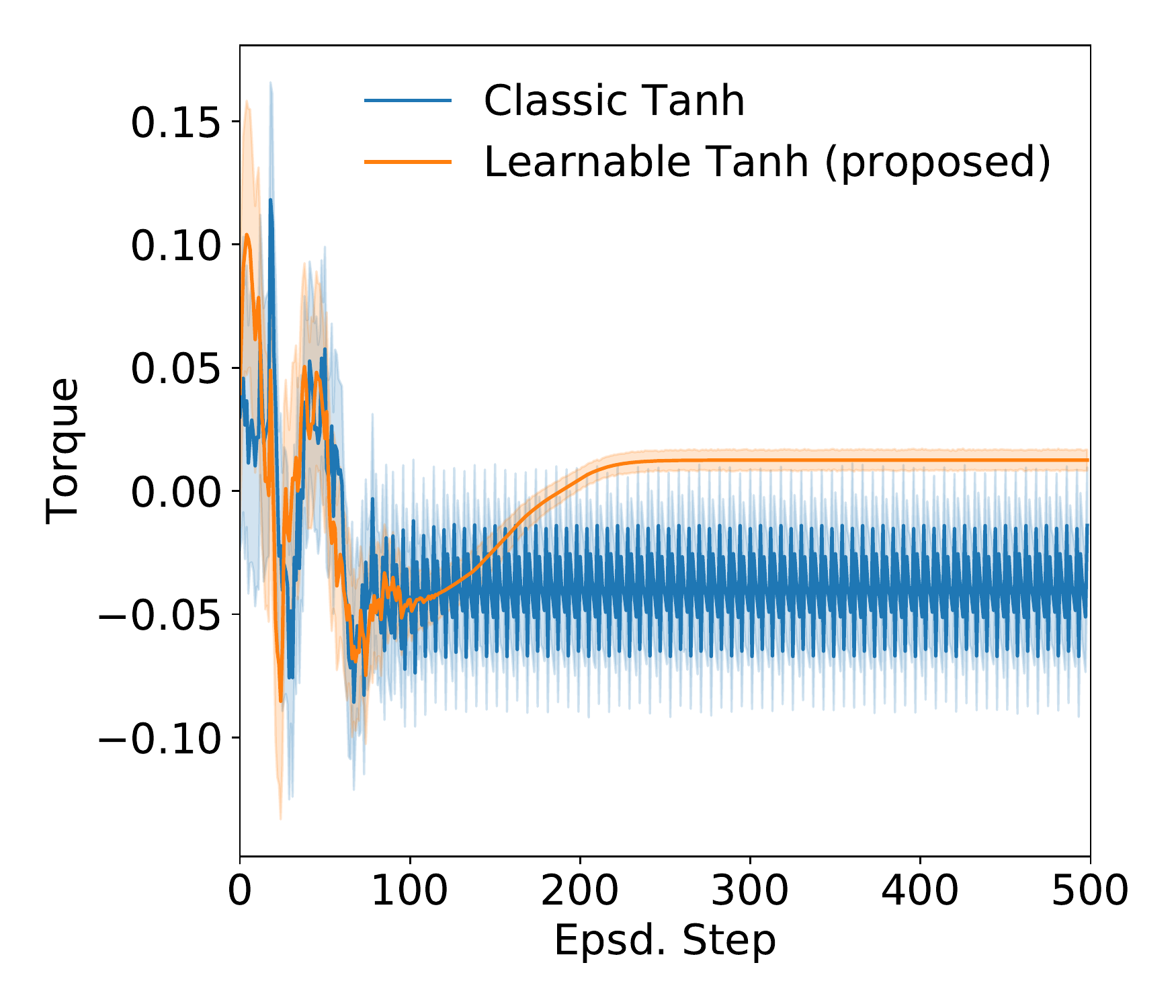}}
    \caption{Inverted pendulum results. No significant improvement in training performance (a). The best models from both methods reported similar reward progression patterns (b, c, d). The proposed method achieves a more stable control where as the classic method oscillates actions to maintain control (e).}
    \label{fig:inv}
\end{figure}

In the inverted pendulum problem (Fig.~\ref{fig:inv}), the improvement is insignificant because the environment featured only one actuator. However, the policy adapted by the proposed agent features a fine balance of actuation signals. In contrast, the classical MLP/Tanh model exerts additional oscillating actuation signals to maintain the achieved optimal state, as shown in Fig.~\ref{fig:inv_act}. This oscillation, imposes a wear and tear challenge on mechanical systems and fatigue risks in biomechanical systems. While this difference is reflected with minimal difference in the environment reward, it is often a critical decision to make in practical applications. 


\subsection{Lunar Lander Results}
\begin{figure}
    \centering
    \subfigure[Training Performance]{
    \centering
        \includegraphics[width=.96\linewidth]{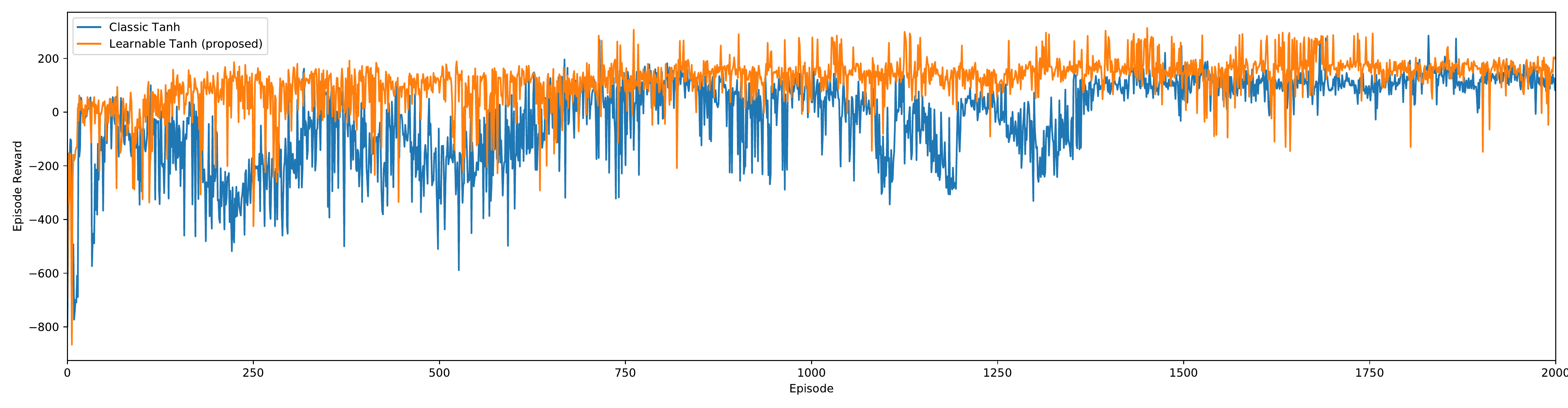}
    }\\
        \subfigure[Step Reward]{
        \label{fig:llc_stpR}
        \centering
            \includegraphics[width=.48\linewidth]{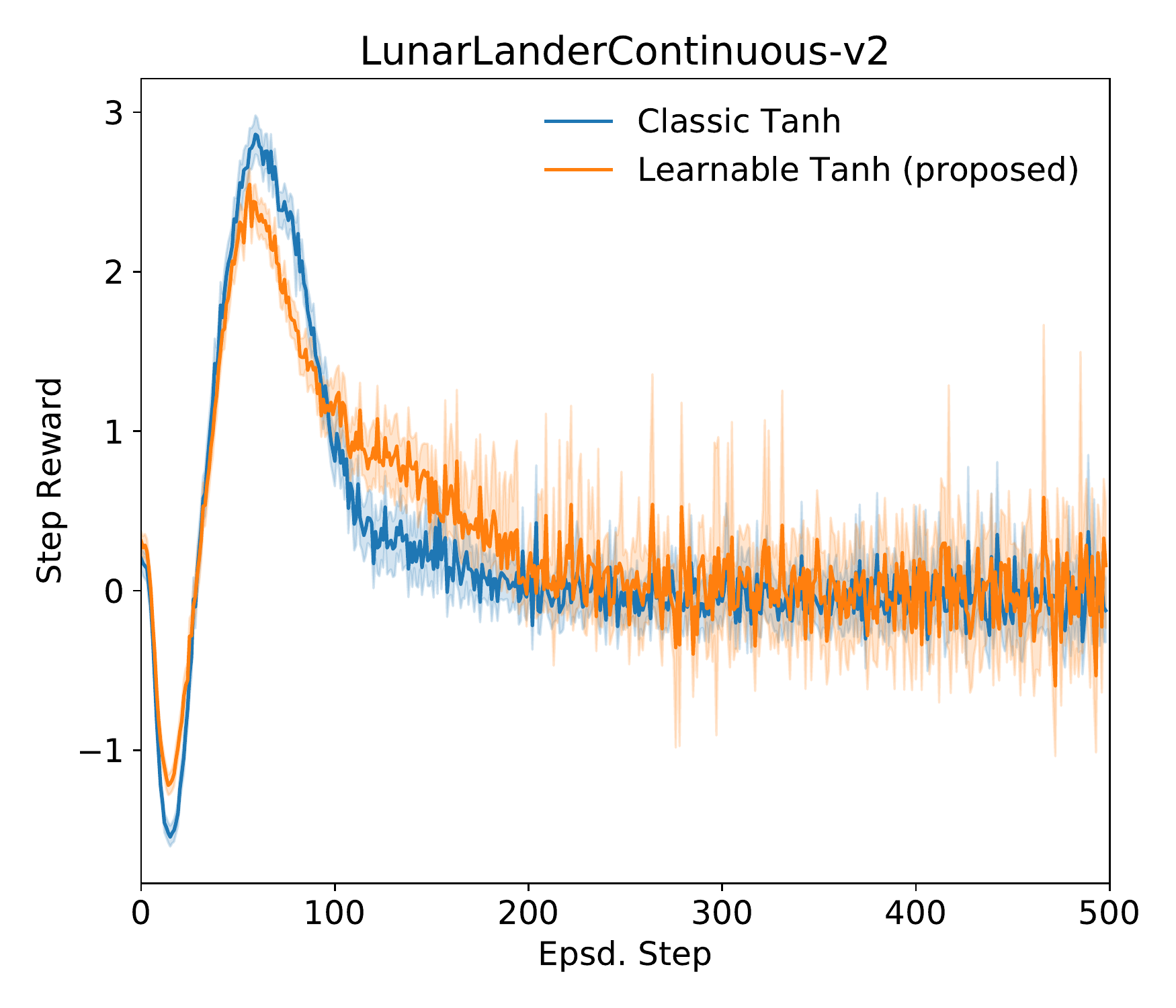}}
        \subfigure[Epsd. Reward]{
        \label{fig:llc_epsR}
        \centering
            \includegraphics[width=.48\linewidth]{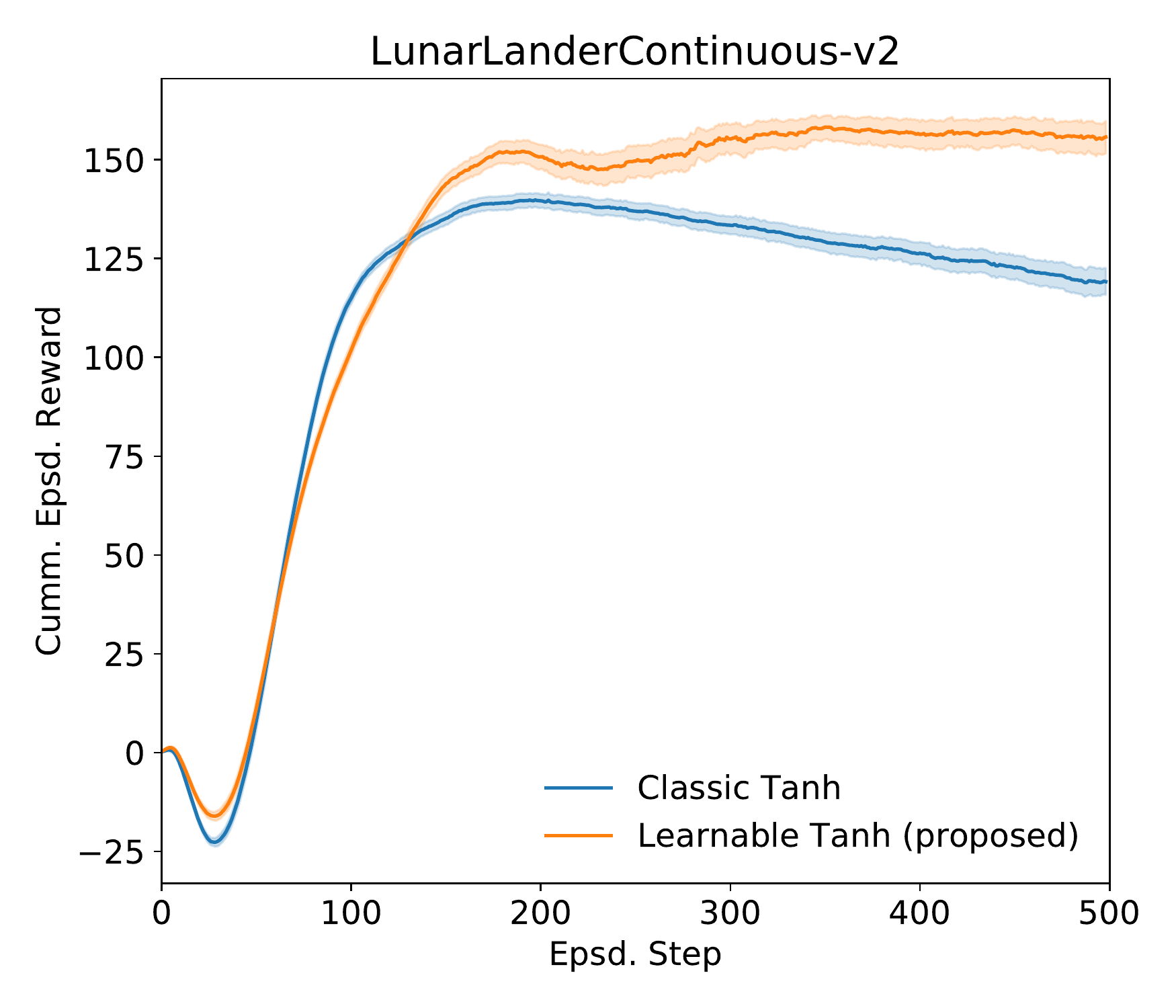}}

    \caption{Lunar lander results. There is a significant improvement in training performance at the early stages of the training (a). Agents with the proposed method outperforms the classical method in terms of reward progression (b and c).}
    \label{fig:llc}
\end{figure}

\begin{figure*}
    \centering
    \subfigure[Landing Trajectory]{
    \label{fig:llc_trj}
    \centering
        \includegraphics[width=1\linewidth]{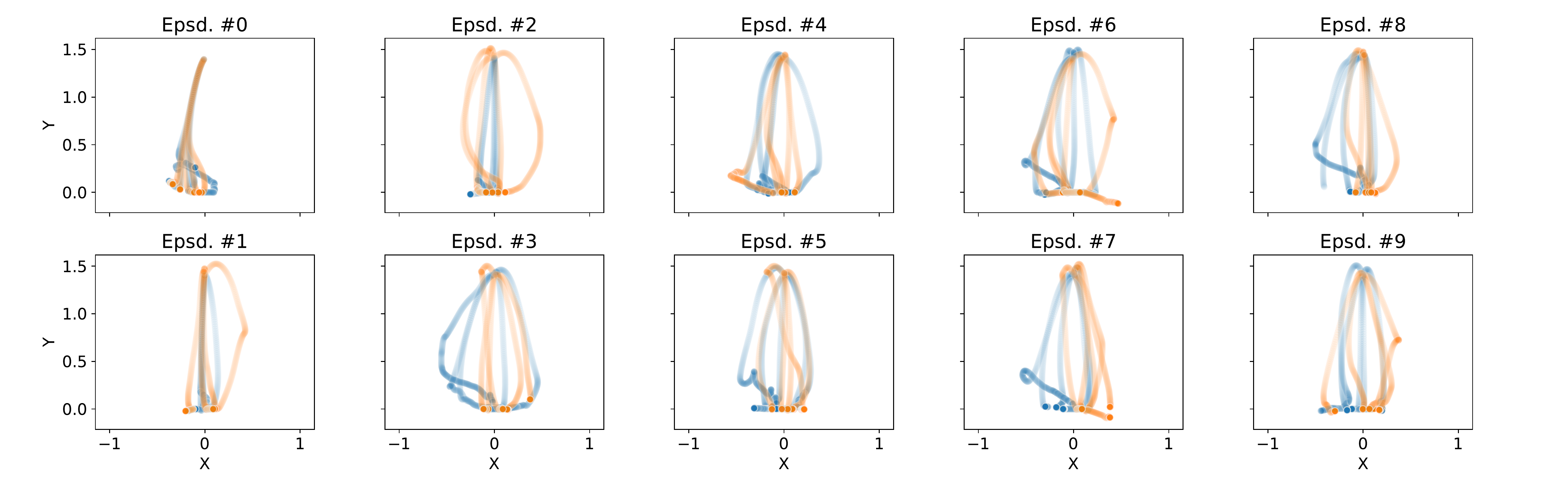}
    }\\

        \subfigure[Position]{
        \label{fig:llc_act}
        \centering
            \includegraphics[width=.31\linewidth]{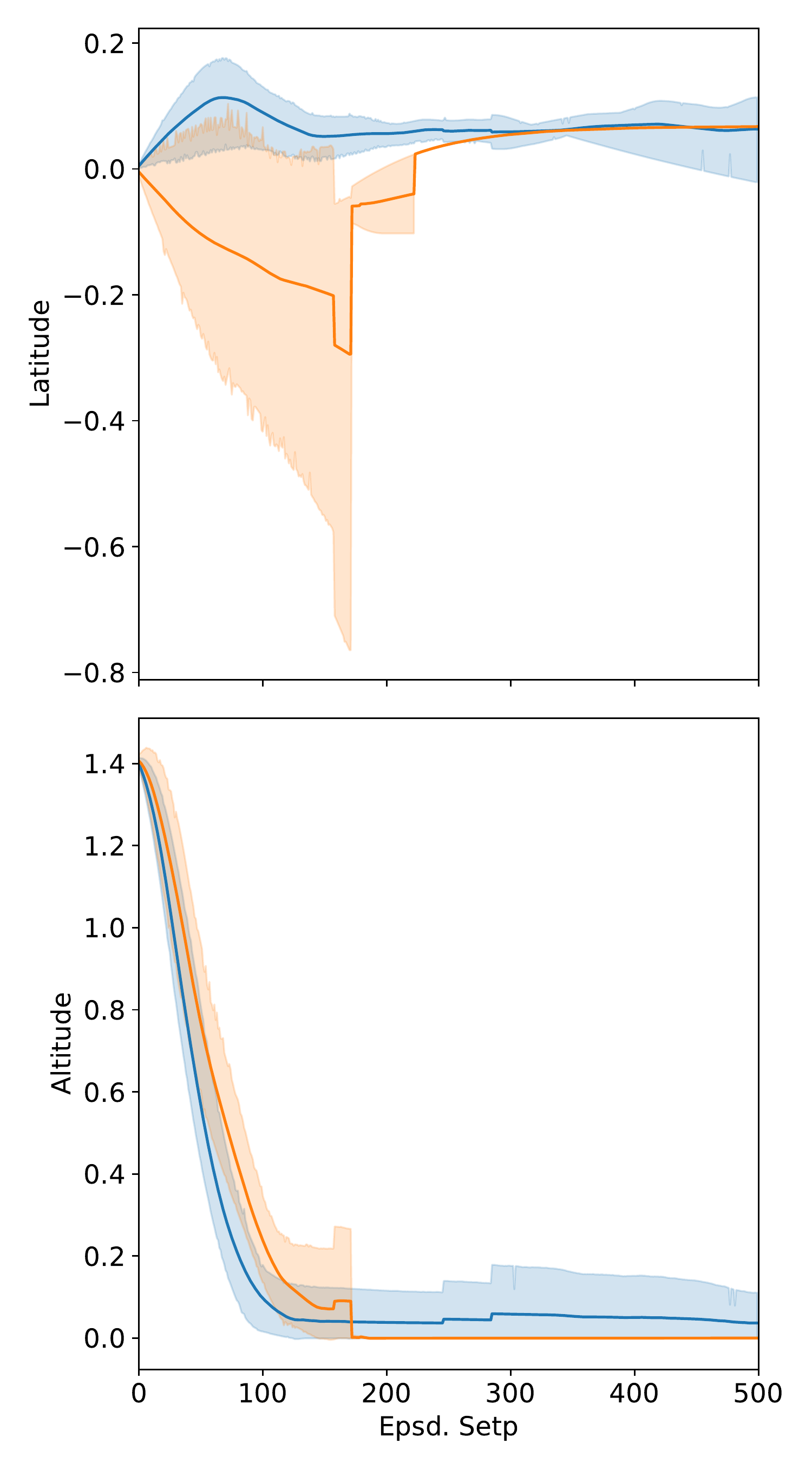}
        }
        \subfigure[Actions]{
        \label{fig:llc_act}
        \centering
            \includegraphics[width=.31\linewidth]{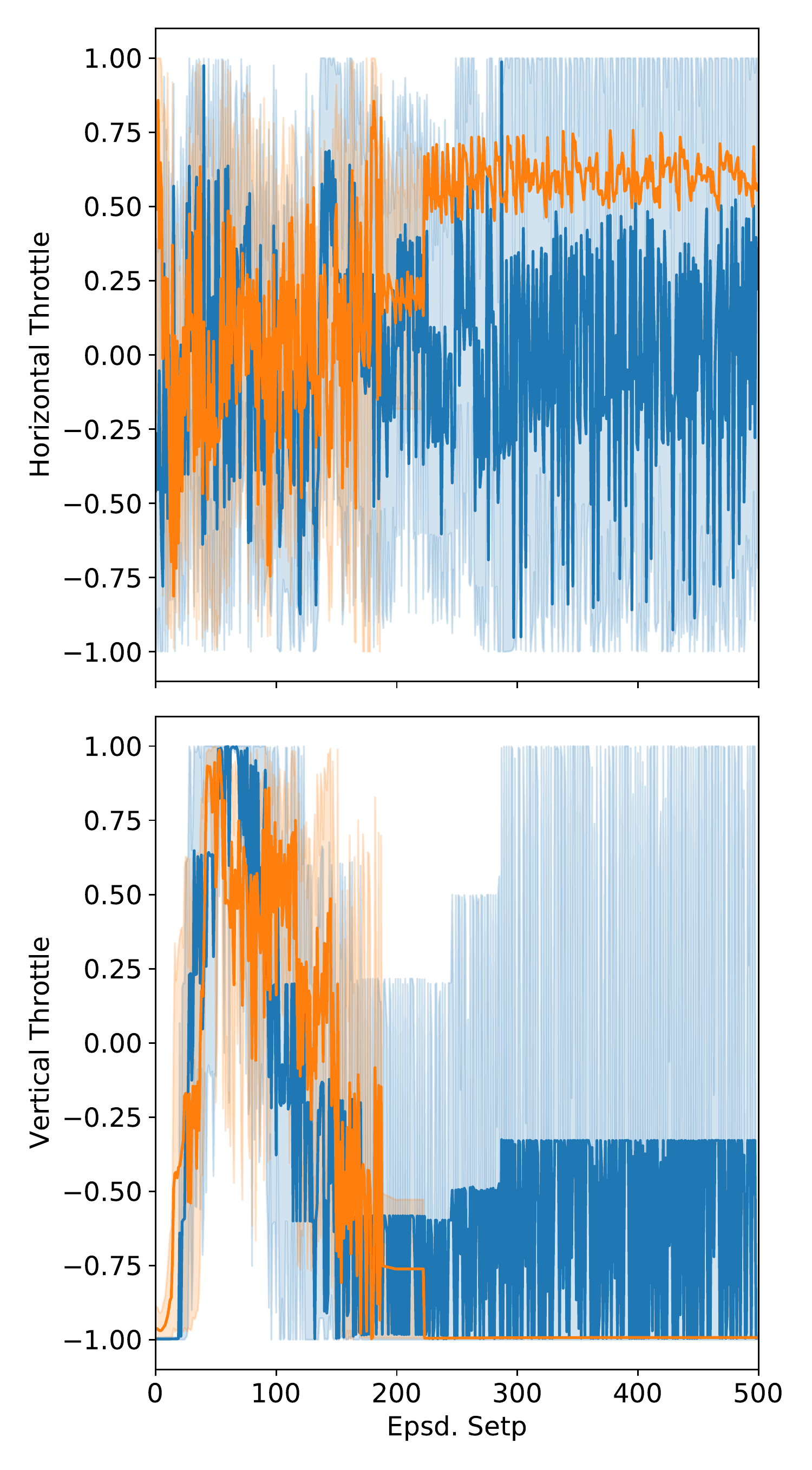}
        }
        \subfigure[Velocity]{
        \label{fig:llc_act}
        \centering
            \includegraphics[width=.31\linewidth]{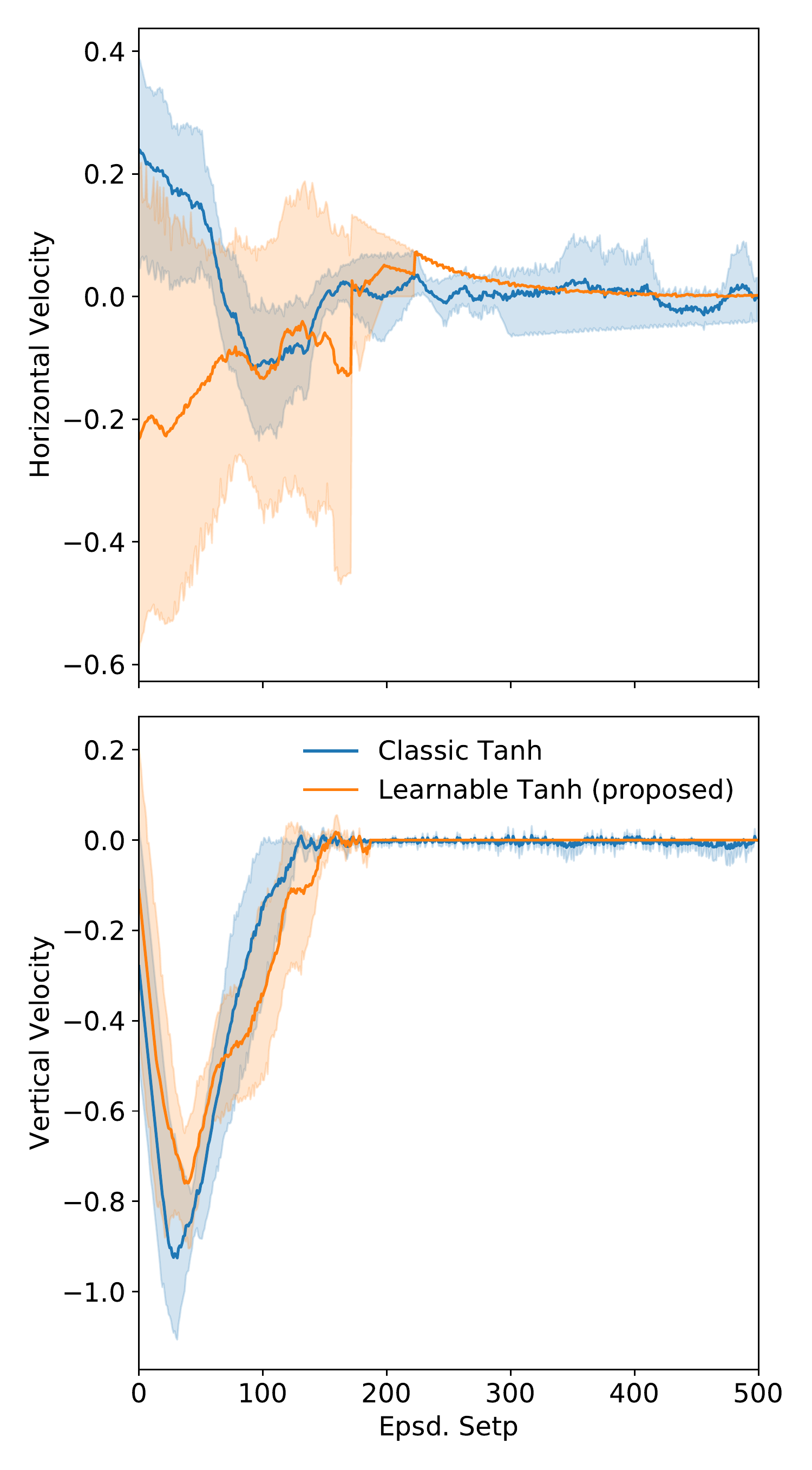}
        }

    \caption{Lunar lander results. Proposed and classical actors adopt different landing trajectories (a). Actors without the proposed method preserves effort by engaging in free falling to a safe altitude (b-bottom) and then exerts more effort to perform safe landing (c-bottom). Actors with the proposed method decelerates and engage in manoeuvring to a safe landing (b, c, d).}
    \label{fig:llctraj}
\end{figure*}

Figure~\ref{fig:llc} shows the training and reward curves of the lunar landing problem. The instant reward curve of the lunar landing problem demonstrate an interesting behaviour in the first $100$ steps. The classic method adopts an energy-conservative policy by shutting down the main throttle and engaging in free falling for $25$ steps to a safe margin and then keep hovering above ground to figure out a safe landing. The conserved energy contributes to the overall reward at each time step. While this allows for faster reward accumulation, this policy becomes less effective with different initial conditions. Depending on the speed and the attack angle, the free-falling policy requires additional effort for manoeuvring the vehicle to the landing pad. The proposed agent, on the other hand, accommodates the initial conditions and slows down the vehicle in the opposite direction to the entry angle to maintain a stable orientation and thus allows for a smoother lateral steering towards a safe landing as shown in Fig~\ref{fig:llctraj} (a and c). It is worth noting that both agents did not perform any type of reasoning or planning. The main difference is the additional degrees of freedom the proposed parametrised activation function offers. These degrees of freedom allow the actor neural network to adopt different response functions to accommodate the initial conditions.

\subsection{Bipedal Walker Results}
\begin{figure*}
    \centering
    \subfigure[Training Performance]{
    \label{fig:bpw_trj}
    \centering
        \includegraphics[width=.96\linewidth]{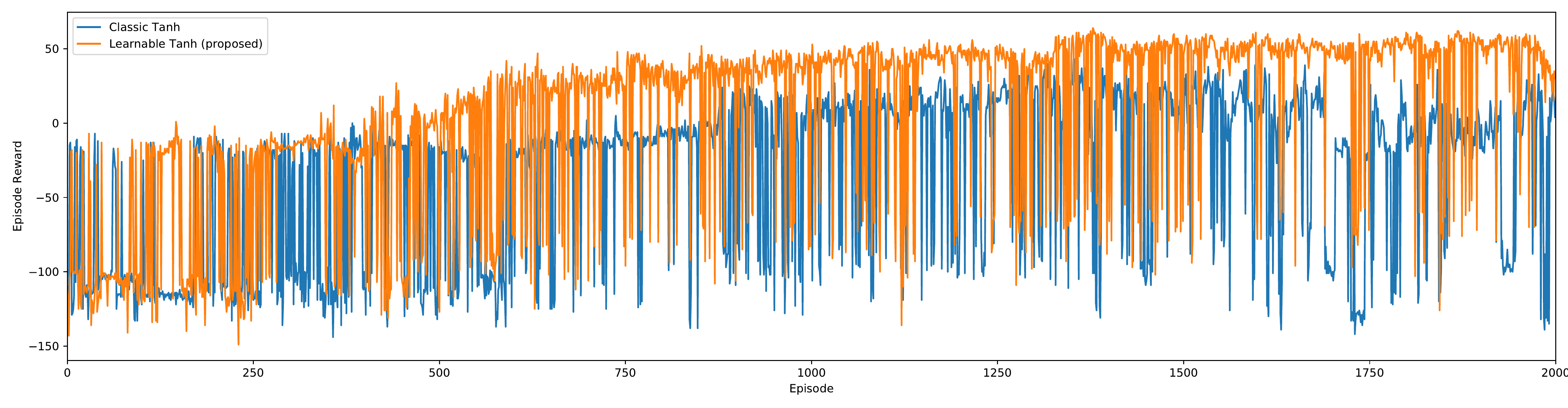}
    }\\

    \begin{minipage}{.31\textwidth}
    \subfigure[Step Reward]{
    \label{fig:bpw_act}
    \centering
        \includegraphics[width=1\linewidth]{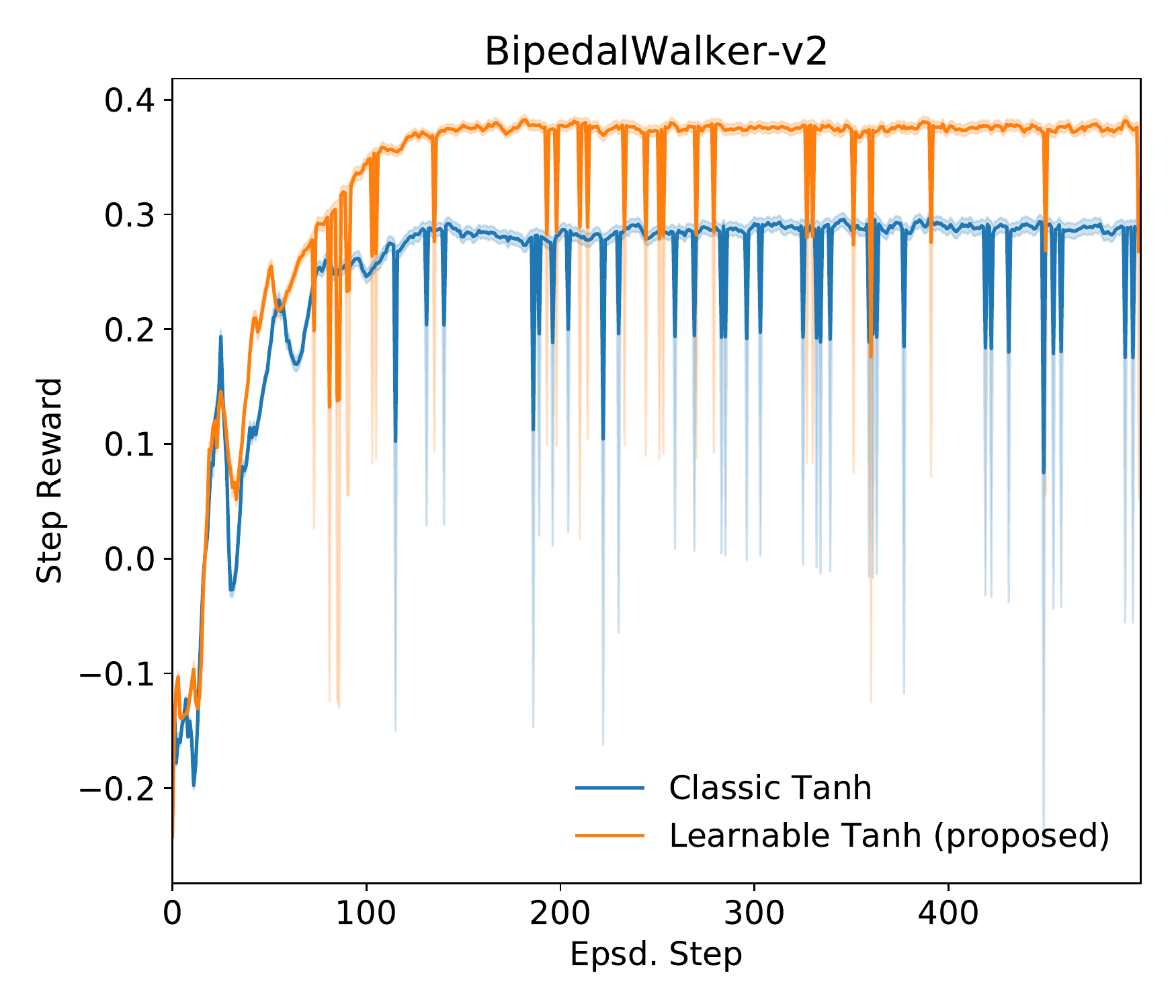}
    }\\
    \subfigure[Epsd. Reward]{
    \label{fig:bpw_act}
    \centering
        \includegraphics[width=1\linewidth]{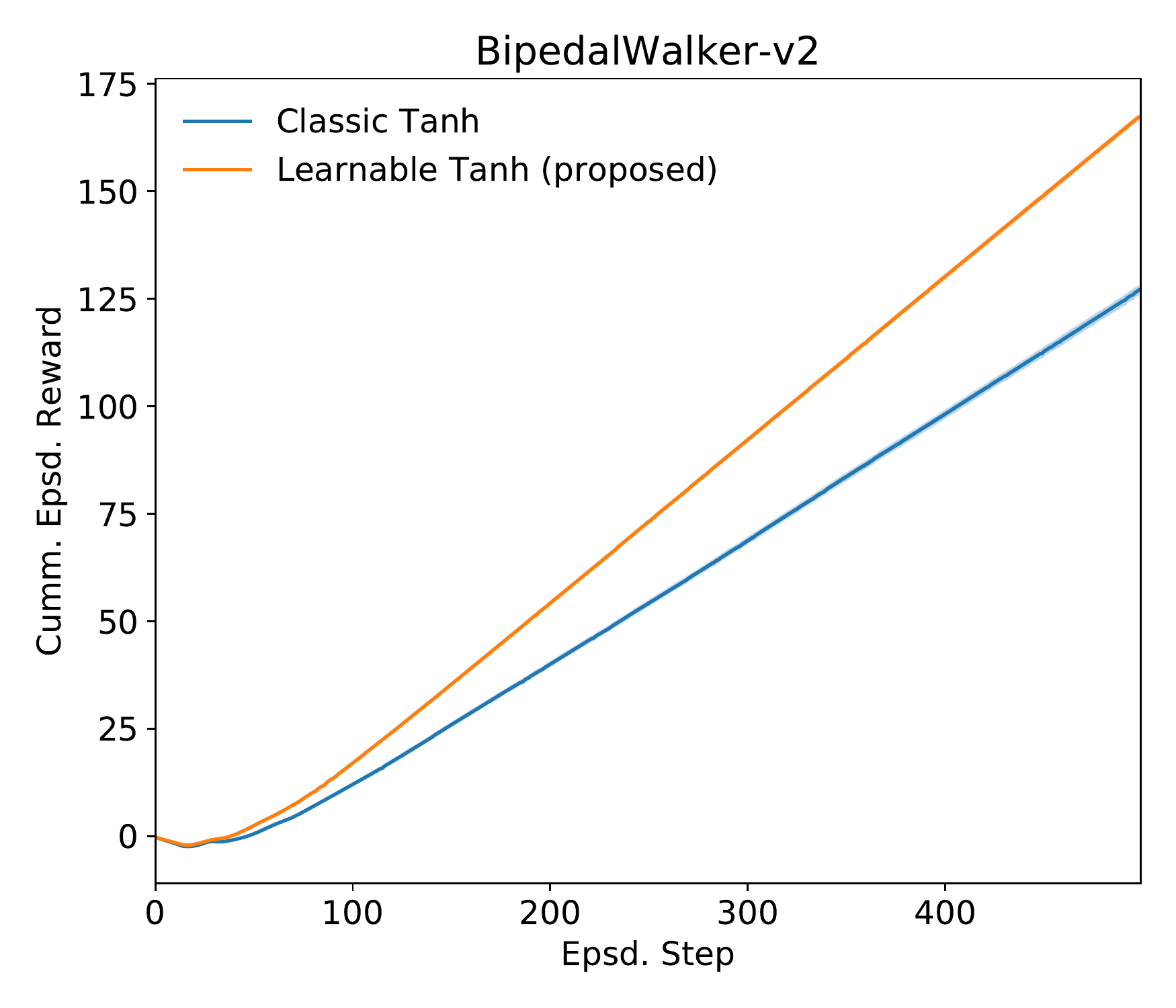}
    }
    \end{minipage}%
    \begin{minipage}{.32\textwidth}
    \subfigure[Observations]{
    \label{fig:bpw_act}
    \centering
        \includegraphics[width=1\linewidth]{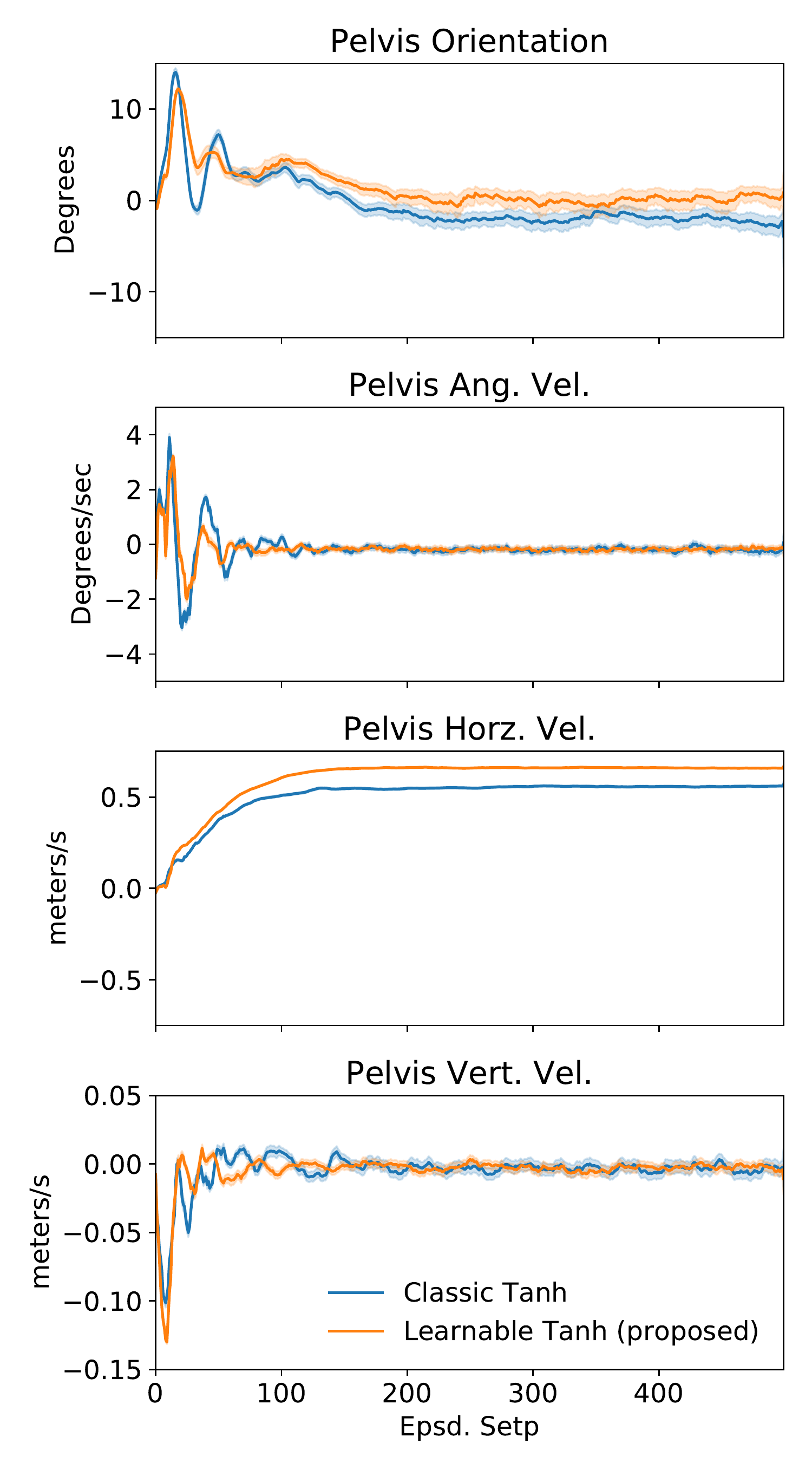}
    }
    \end{minipage}%
    \begin{minipage}{.32\textwidth}
    \subfigure[Actions]{
    \label{fig:bpw_act}
    \centering
        \includegraphics[width=1\linewidth]{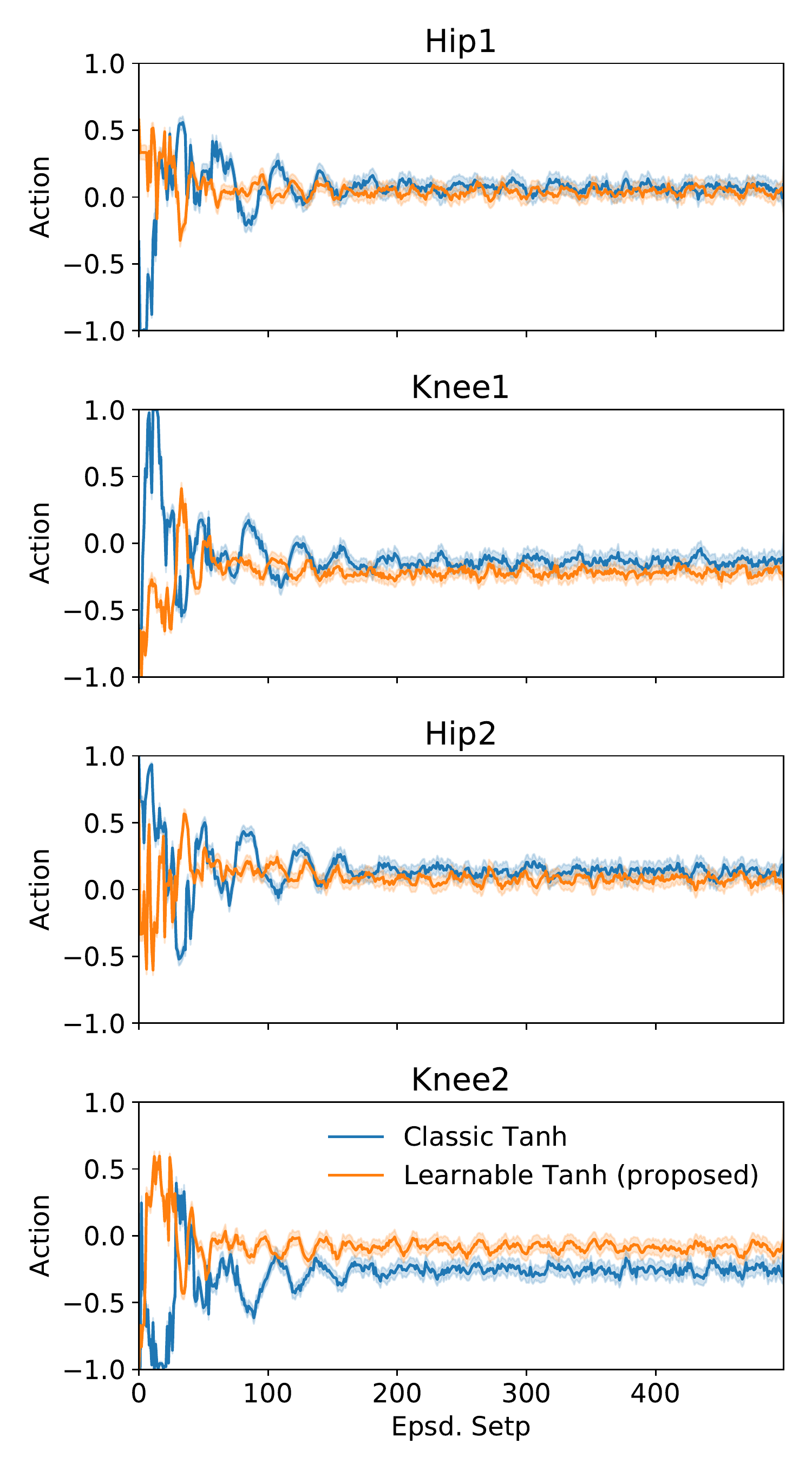}
    }
    \end{minipage}
    \caption{Bipedal walker results. There is a significant improvement in training performance (a). Actors with the proposed method outperforms the classical method in terms of reward progression (b and c), more stable (d), performs the task faster (d) with minimal effort (e).}
    \label{fig:bpw}
\end{figure*}

Training and reward curves of the bipedal walking problem are illustrated in Fig.~\ref{fig:bpw}. In general, the agent with the proposed action mapping out performs the classical agent in the training, step and episode reward curves as shown in Fig.~\ref{fig:bpw}(a, b, c). The spikes in the step reward curves shows instances where agents lost stability and failed to complete the task. The episode reward curve shows that the proposed method allows the agent to run and complete the task faster. This is due to a better coordination between the left and right legs while taking advantage of the gravity to minimise the effort. This is demonstrated in Fig.~\ref{fig:bpw}-d where the proposed agent maintains a pelvis orientation angular velocity and vertical velocity close to zero. This, in return, dedicates the majority of the spent effort towards moving forward. This is also reflected in Fig.~\ref{fig:bpw}-e where the actuation of the proposed agent stabilises faster around zero and thus exploiting the gravity force. In contrast, the classical agent, spends more effort to balance the pelvis and thus it takes longer time to stabilise actuation. Finally, the locomotion actuation patterns in Fig.~\ref{fig:bpw}-e demonstrate the difference between the adapted policies. The classical agent relies more on locomoting using \texttt{Knee2} while the proposed agent provides more synergy between joint actuators. This difference in exploiting the gravity during locomotion is an essential key in successful bipedal locomotion as ``controlled falling'' \citep{novacheck1998biomechanics}.


\section{Conclusions}
\label{sc:conc}
In this paper, we discussed the advantages of adding parameterisation degrees of freedom to the actor in the DDPG actor-critic agent. The proposed methods is simple and straight forward, yet it outperforms the classical actor which utilises the standard $\tanh$ activation function. The main advantage of the proposed method lies in producing stable actuation signals as demonstrated in the inverted pendulum and bipedal walker problems. Another advantage that was apparent in the lunar landing problem is the ability to accommodate changes in initial conditions. This research highlights the importance of parameterised activation functions. While the discussed advantage may be minimal for the majority of supervised learning problems, they are essential for dynamic problems addressed by reinforcement learning. This is because reinforcement learning methods, especially the off-policy ones, rely on previous experiences during training. 

\smallbreak
The advantage of the proposed method in the bipedal walking problem and the wide variety of activation functions demonstrated in Fig.~\ref{fig:res_tanh} suggests a promising potential for solving several biomechanics problems where different muscles have different characteristics and response functions. Applications such as fall detection and prevention~\citep{abobakr2017fall}, ocular motility and the associated cognitive load and motion sickness~\citep{iskander2018ocular, iskander2018review, iskander2019car, Attia2018766}, as well as intent prediction of pedestrians and cyclists~\citep{Saleh2018414,Saleh2020317}. The stability of the training using the parameterised $\tanh$ in an actor-critic architecture also shows potential for advancing the Generative Adversarial Networks (GANs) research for image synthesis~\citep{attia2018realistic}. 

\smallbreak
This research can be expanded in several directions. First, the parameterisation of $\tanh$ can be extended from being deterministic (presented in this paper), to a stochastic parameterisation by inferring the distributions of $k$ and $x_0$. Second, the separation between the policy and the action parts of the actor neural network allows preserving the policy part while fine tuning only the action part to accommodate actuator characterisation discrepancies due to wear and tear during operations. Finally, the modular characterisation of different parts of the actor neural network into observer, policy and action parts requires investigating scheduled training to lock and unlock both parts alternatively to maximise the dedicated function each part the actor carries out.

\end{document}